\documentclass[letterpaper]{article} 
\usepackage[preprint]{aaai2027}  
\usepackage[hyphens]{url}  
\usepackage{graphicx} 
\usepackage{natbib}  
\usepackage{caption} 
\usepackage{algorithm}
\usepackage{algorithmic}

\usepackage{newfloat}
\usepackage{listings}
\DeclareCaptionStyle{ruled}{labelfont=normalfont,labelsep=colon,strut=off} 
\floatstyle{ruled}
\newfloat{listing}{tb}{lst}{}
\floatname{listing}{Listing}

\usepackage{booktabs}

\usepackage{amsmath}

\usepackage{cleveref}
\usepackage{xcolor}

\usepackage[table]{xcolor}

\usepackage{multirow}

\usepackage{pifont}

\title{CMT-RAG: Complementary Memory Traces for \\ Multi-turn Multi-hop RAG}
\author{
    Lang Zhou\textsuperscript{1,2}, 
    \textbf{Yingjian Chen}\textsuperscript{1,2},
    \textbf{Shuxuan Li}\textsuperscript{2},
    \textbf{Kun-Yu Lin}\textsuperscript{3},
    \textbf{Zhilin Zhao}\textsuperscript{1,2}
}
\affiliations{
    \textsuperscript{\rm 1}Sun Yat-sen University \\
    \textsuperscript{\rm 2}Shenzhen Loop Area Institude \\
    \textsuperscript{\rm 3}The University of Hong Kong \\
}

\begin{document}

\maketitle

\begin{abstract}
Multi-turn information-seeking conversations require both multi-hop reasoning and long-range dependency tracking across turns. However, existing RAG systems typically represent conversational memory as raw dialogue history, rewritten queries, or unstructured summaries, making it difficult to recover the specific prior reasoning steps and evidence required for follow-up queries. Our key insight is to align conversational memory with retrieval by representing dialogue context as sub-question-level reasoning traces. Building on this insight, we introduce \textbf{MuMu-QA}, a benchmark for multi-turn multi-hop RAG with explicit cross-turn sub-question dependency annotations, and \textbf{CMT-RAG}, a complementary memory framework for this setting. At each turn, CMT-RAG employs a state-space trace generator, whose recurrent state serves as runtime memory, to incorporate recent conversational context and decompose the current query into structured trace drafts containing retrieval-oriented sub-questions and dependencies on earlier traces. It then grounds these drafts with retrieved evidence and stores them as persistent memory traces in a session-level DAG, enabling future turns to efficiently recover relevant prior reasoning and evidence. Experiments on MuMu-QA and corpus-level RAG benchmarks show that CMT-RAG consistently outperforms five categories of RAG baselines in answer accuracy.
\end{abstract}

\section{Introduction}

\begin{figure}[t]
  \centering
  \includegraphics[width=\columnwidth]{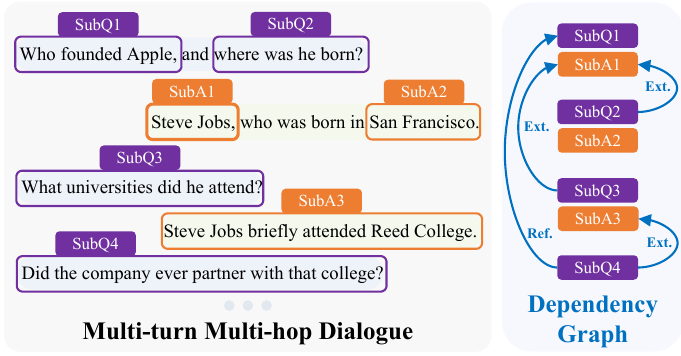}
  \caption{A multi-turn multi-hop conversation with cross-turn dependencies. The graph illustrates two dependency types: \textbf{Ext.} (Predicate Extension) queries a new attribute or relation of a previously resolved target, and \textbf{Ref.} (Entity Reference) directly reuses a previously introduced entity.}
  \label{fig-overview}
\end{figure}

Retrieval-augmented generation (RAG) is increasingly deployed in extended information-seeking conversations, where users refine questions, omit repeated entities, and build new requests on earlier answers~\cite{ye2026hmem,laban2026llms,hu2025memorybench}. In such settings, a new turn often remains context-dependent while requiring multi-hop evidence seeking, since answering it may require decomposing the query into several sub-questions whose dependencies span earlier turns. \Cref{fig-overview} shows a representative case. The system must retrieve evidence for the current turn and identify the specific prior reasoning step whose subject or entity is being extended or reused. We refer to this setting as \emph{multi-turn multi-hop conversational RAG with complementary sub-question dependencies}.

This setting exposes a mismatch between how conversational RAG stores memory and how retrieval actually operates. Query rewriting converts a context-dependent turn into a standalone query~\cite{anantha2021qrecc,mo2023convgqr,zhu2025convsearchr1}, which is effective for local coreference but compresses dependency chains into a single query, obscuring intermediate retrieval targets. Query decomposition exposes sub-question structure for multi-hop retrieval~\cite{trivedi2023ircot,khot2023decomp,chen2026logicrag,ye2025qdream}, yet typically assumes a self-contained query with dependencies confined to the current turn. Memory-based conversational systems store histories, summaries, or embeddings~\cite{liu2024chatqa,zhong2024memorybank}, while dialogue graphs model utterance-level relations~\cite{li2020molweni,fan2023rltst,zhu2025cidgraphrag}. None explicitly represents retrieval-level dependencies across turns. As a result, retrievers require sub-question-level memory, whereas existing systems largely maintain only turn-level context.

Our key insight is to align conversational memory with retrieval by storing dialogue context as sub-question-level reasoning traces. A useful memory unit for this setting should preserve the retrieval target, expose the dependency that resolves missing arguments, and retain the evidence that made the earlier answer valid. A sub-question-level trace provides this unit by packaging a past reasoning step as an addressable object. When a later turn depends on it, the system can recover the relevant trace through its dependency links and keywords, then reuse the associated evidence under the current query. Cross-turn recall is therefore reduced from global history interpretation to trace selection and evidence reuse.

Accordingly, we propose \textbf{CMT-RAG}, a framework built around \emph{complementary memory traces}. At each turn, a state-space trace generator consumes the current query and its recurrent state to produce structured \emph{trace drafts}, each containing a sub-question, trace keywords, and dependencies on earlier traces. After answer-reference resolution, fresh evidence is retrieved for each sub-question, and the draft is completed by attaching the corresponding paragraph identifiers. Dependency links access prerequisite DAG nodes, while trace keywords retrieve an additional historical trace through lexical matching. The reader transiently combines the accessed historical evidence with the freshly retrieved evidence. After answering, the completed trace and its sub-answer are appended to the DAG. The recurrent state thus maintains local discourse continuity, while the trace DAG preserves explicit long-range dependencies and reusable evidence. The downstream reader remains stateless, receiving only the resolved sub-question and its assembled evidence.

To study this problem directly, we introduce \textbf{MuMu-QA}, a benchmark that reorganizes multi-hop questions into multi-turn dialogues with cross-turn dependency annotations. Existing multi-turn RAG benchmarks evaluate conversational retrieval and generation at the turn level, without exposing which current sub-question depends on which prior sub-question. MuMu-QA fills this gap by providing dialogue-wide sub-question identifiers, trace keywords, dependency edges, supporting paragraph IDs, and full trace DAG supervision, with long-dialogue splits for stress-testing dependency recovery beyond short history replay. Experiments on MuMu-QA show that CMT-RAG achieves the best answer accuracy among direct C-RAG, query rewriting, agentic retrieval, decomposition-based RAG, and dialogue-structure baselines. With a stateless Qwen3-32B reader, it reaches 41.73 EM and 55.63 F1 using top-5 retrieval while keeping cross-turn memory outside the reader.

\section{Preliminaries}

This section fixes the notation and evaluation target used throughout the paper. We first define multi-turn multi-hop conversational RAG as trace-DAG induction, where each trace is a retrieval-level memory unit that binds a sub-question to dependencies, keywords, and evidence. We then describe MuMu-QA as the benchmark instantiation of this formulation, with supervision over sub-question dependencies and reusable paragraph evidence.

\subsection{Task Formalization}
\label{sec:task-formalization}

We consider a multi-turn multi-hop conversational RAG session over an unstructured corpus $\mathcal{C}$. The dialogue is an ordered sequence of $T$ user turns,
\begin{equation}
\mathcal{D} = \langle q_1, q_2, \ldots, q_T\rangle,
\end{equation}
where $q_t$ denotes the user query at turn $t$. For each turn, the system must produce an answer $a_t$ grounded in evidence from $\mathcal{C}$. The distinctive difficulty is that a turn may contain multiple retrieval-relevant sub-questions, each of which may depend on information from earlier turns. We formalize this setting as session-level induction of a directed acyclic graph of \emph{traces},
\begin{equation}
\mathcal{G} = (\mathcal{V}, \mathcal{E}_{\mathcal{G}}),
\end{equation}
where each node $(\mathcal{T}_k,a_k) \in \mathcal{V}$ consists of a trace and its answer. The trace $\mathcal{T}_k$ binds a sub-question, trace keywords, dependency edges, and retrieved paragraph identifiers as
\begin{equation}
\mathcal{T}_k = \bigl(q_k^{\text{sub}},\ kw_k,\ \mathrm{deps}(k),\ \mathrm{para\_ids}_k\bigr),
\end{equation}
where $q_k^{\text{sub}}$ is the natural-language sub-question consumed by the reader and retriever, $kw_k$ are lookup anchors for the trace DAG, $\mathrm{deps}(k) \subseteq \{1,\ldots,k-1\}$ lists prerequisite traces, and $\mathrm{para\_ids}_k$ identifies the paragraphs directly retrieved from $\mathcal{C}$ for this sub-question. Each edge $(\mathcal{T}_j,\mathcal{T}_i) \in \mathcal{E}_{\mathcal{G}}$ states that trace $\mathcal{T}_i$ relies on the entity or subject introduced by trace $\mathcal{T}_j$.

At turn $t$, the trace generator maps the current query $q_t$ and recurrent state $h_{t-1}$ to an ordered set of trace drafts. CMT-RAG then resolves their predicted dependencies against $\mathcal{G}_{<t}$ and appends the completed traces as $\Delta\mathcal{G}_t$. This formulation couples two structured operations. The system must decompose the current turn into retrieval units and link those units to prior traces whose subjects or entities remain necessary. The target memory unit is not a whole utterance or an unstructured history summary. It is a trace whose fields are directly consumed by retrieval, DAG lookup, and answering.

\subsection{Benchmark Construction}
\label{sec:mumu-qa-overview}
MuMu-QA instantiates this formulation as a benchmark for multi-turn multi-hop RAG. Existing multi-turn C-RAG benchmarks supervise standalone-query rewriting or turn-level answers~\cite{ali2026recor, cheng2025coral, katsis2025mtrag}, without annotating dependencies at sub-question granularity. We construct \textbf{MuMu-QA} from MuSiQue~\cite{trivedi2022musique}, using its sub-question decompositions, intermediate answers, and supporting paragraphs to derive supervision for trace generation.

\begin{figure}[t]
  \centering
  \includegraphics[width=0.85\columnwidth]{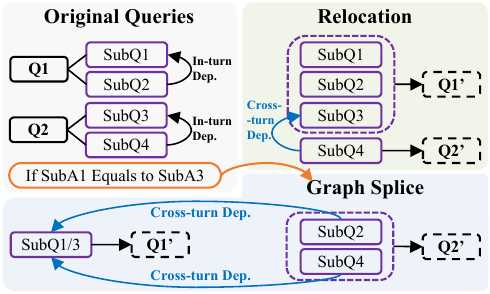}
  \caption{MuMu-QA synthesis operators. Parent questions are decomposed into sub-questions with in-turn dependencies. \textbf{Sub-question Relocation} moves a sub-question to a later turn to create cross-turn dependencies, while \textbf{Graph Splicing} joins two reasoning chains via a bridge answer.}
  \label{fig:data-synthesis}
\end{figure}

\begin{figure*}[t]
  \centering
  \includegraphics[width=0.95\linewidth]{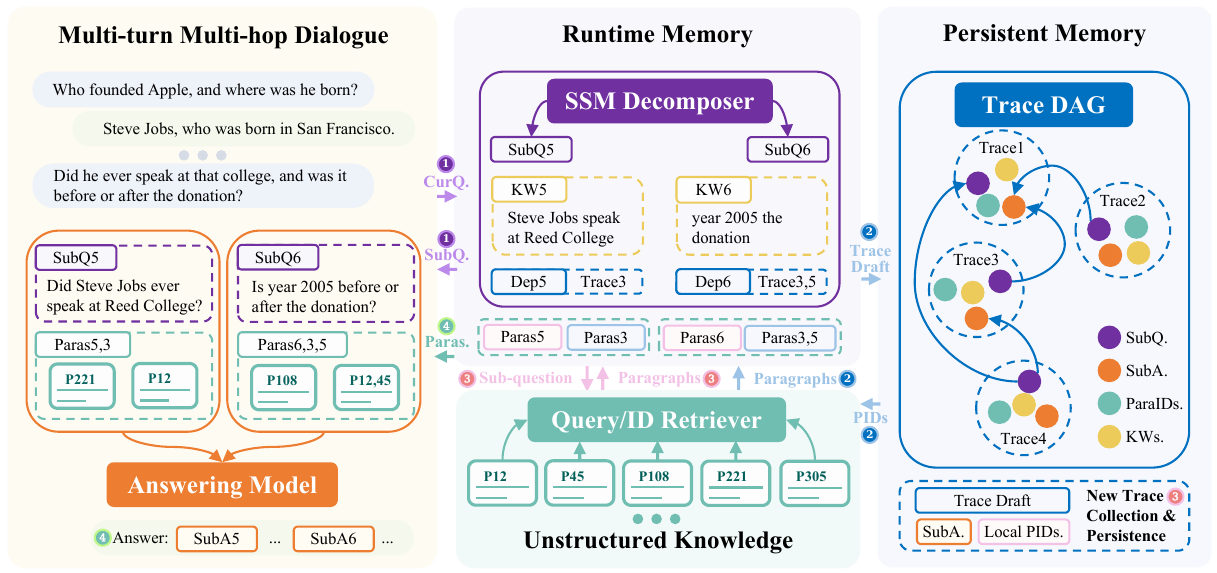}
  \caption{Overview of CMT-RAG. The framework consists of four stages: (i) Trace generation, where a state-space model (SSM) maintains a recurrent state to generate structured trace drafts; (ii) Reference resolution, where cross-turn dependencies are resolved through the trace DAG; (iii) Evidence retrieval and trace update, where supporting paragraphs are retrieved and completed traces are written back to the DAG; and (iv) Question answering, where the stateless reader answers the resolved sub-questions using only the retrieved evidence, without replaying the dialogue history, before producing the final response.}
  \label{fig:cmt-framework}
\end{figure*}

As illustrated in \Cref{fig:data-synthesis}, dialogues are synthesized using two operators. \textit{Sub-question Relocation} moves a sub-question from a multi-hop question into a separate turn and rewrites the remaining question as a follow-up that depends on the relocated answer. \textit{Graph Splicing} links two reasoning chains through a shared bridge answer: a seed turn first resolves the bridge entity, and a later turn continues reasoning from that entity with explicit dependencies on earlier trace nodes. Together, these operators generate short dialogues with controlled cross-turn dependencies. We further construct long dialogues by interleaving topic-related sessions, producing conversations of up to several dozen turns for training and evaluation. Full construction details are provided in Appendix~\ref{app:mumu-qa}.
\section{Method}

CMT-RAG instantiates the trace-DAG formulation with two complementary memory channels: a state-space trace generator that captures local discourse to produce structured trace drafts, and a session-level trace DAG that persistently stores traces for dependency-aware retrieval and evidence reuse. As shown in \Cref{fig:cmt-framework}, this design externalizes conversational state from the answering model, which remains stateless while the recurrent state and trace DAG jointly maintain local and long-range conversational memory.

\subsection{State-Space Trace Generation}
\label{sec:ssm-decomposer}
We instantiate the trace generator with a state space model (SSM) backbone based on Mamba-2~\cite{dao2024mamba2, gu2024mamba} while preserving its selective state-space mixer, whose recurrent state serves as a compact carrier of local discourse context. This enables the model to avoid repeatedly encoding the full dialogue history at each turn, thereby reducing exposure to lost-in-the-middle effects~\cite{yu2025mitigate, liu2024lostmiddle}. We further adapt the backbone through Low-Rank Adaptation (LoRA) fine-tuning and Direct Preference Optimization (DPO), together with a structured output vocabulary, so that it generates trace drafts rather than free-form plans.

At turn $t$, the generator receives the current query $q_t$, the previous hidden state $h_{t-1}$, and emits a set of draft traces and an updated state,
\begin{equation}
\begin{aligned}
\{\mathcal{T}^{\mathrm{draft}}_k\}_{k \in t}, h_t
&= \textsc{TraceGen}_{\theta}(h_{t-1}, q_t), \\
\end{aligned}
\end{equation}
the updated state $h_t$ carrying local continuity such as topic focus, intent shifts and surface coreference, is passed to the next turn.

\paragraph{Structured trace drafts.}
Each draft has the form
\begin{equation}
\mathcal{T}^{\mathrm{draft}}_k
= \bigl(q_k^{\mathrm{decom}},\ kw_k,\ \mathrm{deps}(k)\bigr),
\end{equation}
where $q_k^{\mathrm{decom}}$ is the decomposed sub-question, $kw_k$ contains trace keywords for DAG lookup, and $\mathrm{deps}(k)$ lists prerequisite trace identifiers. 
We separate keywords from sub-questions into two fields serving different purposes. The sub-question is optimized as natural-language input for the reader and dense retrieval after reference resolution, whereas the keyword field is optimized for efficient trace-DAG lookup via lightweight lexical matching.

\paragraph{Global trace namespace.}
CMT-RAG maintains an append-only namespace shared across the dialogue. Each completed trace is assigned a persistent trace identifier, $\texttt{[T\_1]},\ldots,\texttt{[T\_K]}$, and a corresponding answer reference token $\texttt{[A\_1]},\ldots,\texttt{[A\_K]}$. The generator can therefore explicitly reference prior traces through $\mathrm{deps}(k)$.


\subsection{Trace DAG as Persistent Memory}
\label{sec:trace-completion}
The trace DAG maintains a durable session-level memory. For each draft, CMT-RAG first resolves answer-reference tokens in $q_k^{\mathrm{decom}}$ using the answer attached to each referenced DAG node, and produces the resolved sub-question $q_k^{\mathrm{sub}}$. The system then performs two evidence retrieval operations. Dependency edges directly recover evidence from prerequisite traces, while trace keywords retrieve stored paragraph identifiers from relevant historical traces. Together, they form $\mathcal{P}^{\mathrm{prior}}_k$. Since the keywords are generated as normalized retrieval anchors, lightweight lexical lookup suffices to identify relevant traces without maintaining a separate dense index.

Fresh retrieval is then issued from $q_k^{\mathrm{sub}}$ against the external corpus $\mathcal{C}$ to obtain the paragraph identifier set $\mathcal{P}^{\mathrm{local}}_k$. The assembled paragraph set $\mathcal{P}_k$ is used for the latter inference stage. The completed trace stores the resolved sub-question, keywords, dependency links, and the newly retrieved paragraph set $\mathcal{P}^{\mathrm{local}}_k$,
\begin{equation}
\begin{aligned}
\mathcal{T}_k
&= \bigl(q_k^{\mathrm{sub}},\ kw_k,\ \mathrm{deps}(k),\ \mathcal{P}^{\mathrm{local}}_k\bigr), \\
\mathcal{P}_k
&= \mathcal{P}^{\mathrm{local}}_k \cup \mathcal{P}^{\mathrm{prior}}_k .
\end{aligned}
\end{equation}
After the reader produces answer $a_k$, the completed DAG node $(\mathcal{T}_k, a_k)$ is appended to $\mathcal{G}$ in topological order. The trace and answer are subsequently accessed through different mechanisms: answer references resolve missing arguments in future sub-questions, whereas trace keywords retrieve relevant traces together with their supporting evidence. 

\subsection{Curriculum Learning and DPO Training}
\label{sec:joint-training}

The trace generator is trained with curriculum-based supervised fine-tuning under progressively longer conversational contexts, followed by DPO to align generated traces with downstream retrieval and question answering.

\paragraph{Three-stage curriculum.}
Supervised training follows a three-stage curriculum. Stage~1 trains on single-turn examples to learn the trace syntax and basic decomposition structure. Stage~2 introduces short multi-turn dialogues, and Stage~3 further extends training to longer dialogues. Throughout all stages, the primary objective is the autoregressive language modeling loss $\mathcal{L}_{\mathrm{lm}}$ over the linearized gold trace sequence. For a training instance $x_{1:T}$, 
\begin{equation}
\mathcal{L}_{\mathrm{lm}}
= - \sum_{t=1}^{T} \log p_{\theta}(x_t \mid x_{<t}).
\end{equation} 
Later stages progressively increase dialogue length and cross-turn dependency density, while replaying earlier-stage examples to preserve the basic trace representation.

\paragraph{DPO training.}
The DPO stage samples multiple candidate traces per turn from the long-dialogue model and executes them through the fixed retrieval--reader pipeline. Candidate traces are ranked using a composite reward combining final-answer F1 ($F_{\mathrm{final}}$) and matched sub-question F1 ($F_{\mathrm{sub}}$):
\begin{equation}
\label{eq:dpo-pair-reward}
R(\tau;c)
=
F_{\mathrm{final}}(\tau)
+
\gamma F_{\mathrm{sub}}(\tau),
\end{equation}
\noindent
where, $c$ denotes the current dialogue context for the policy. $F_{\mathrm{final}}$ is computed between the reader's final answer induced by candidate trace $\tau$ and the ground truth, while $F_{\mathrm{sub}}$ averages F1 over lexically matched generated and reference sub-questions. Invalid traces are filtered after pair construction rather than rewarded explicitly. Preference pairs are formed from sufficiently separated high- and low-reward traces, 
and DPO trains the trace generator to assign higher probability to preferred traces than rejected ones under the frozen long-dialogue SFT reference policy.

\subsection{Inference with CMT-RAG}
\label{sec:inference}

\Cref{alg:trace_inference} summarizes inference for a single user turn. Conversational state is fully externalized into the recurrent state and trace DAG, allowing the reader to remain stateless and answer each resolved sub-question using only its assembled evidence. An additional reader call then aggregates the user query together with the current-turn sub-questions and sub-answers into the final answer $a_t$, while the updated DAG is carried forward to subsequent turns.

\begin{algorithm}[t]
\caption{CMT-RAG inference at turn $t$.}
\label{alg:trace_inference}
\begin{algorithmic}[1]
\REQUIRE State $h_{t-1}$, trace DAG $\mathcal{G}_{<t}$, query $q_t$,
retriever $\mathcal{R}$, reader $\mathcal{M}$

\STATE $\{\mathcal{T}_k^{\mathrm{draft}}\}_t,h_t
\gets \textsc{Gen}_{\mathrm{SSM}}(h_{t-1},q_t)$
\STATE $\mathcal{G}_t \gets \mathcal{G}_{<t}$

\FOR{each $\mathcal{T}_k^{\mathrm{draft}}
=(q_k^{\mathrm{decom}},kw_k,\mathrm{deps}_k)$ in topological order}

    \STATE $q_k^{\mathrm{sub}}
    \gets \textsc{RefRes}
    (q_k^{\mathrm{decom}},\mathrm{deps}_k,\mathcal{G}_t)$

    \STATE $\mathcal{T}_k^{\mathrm{prior}}
    \gets \textsc{Index}(\mathcal{G}_t,kw_k)\cup \mathrm{deps}_k$

    \STATE $\mathrm{pids}_k^{\mathrm{prior}}
    \gets \textsc{GetPIDs}(\mathcal{T}_k^{\mathrm{prior}})$

    \STATE $\mathrm{pids}_k^{\mathrm{local}}
    \gets \mathcal{R}(q_k^{\mathrm{sub}})$
    
    \STATE $\mathrm{paras}_k
    \gets \textsc{Get}
    (\mathrm{pids}_k^{\mathrm{prior}}\cup\mathrm{pids}_k^{\mathrm{local}})$

    \STATE $a_k
    \gets \mathcal{M}(q_k^{\mathrm{sub}},\mathrm{paras}_k)$

    \STATE $\mathcal{T}_k
    \gets (q_k^{\mathrm{sub}},kw_k,\mathrm{deps}_k,
    \mathrm{pids}_k^{\mathrm{local}})$

    \STATE $\mathcal{G}_t
    \gets \mathcal{G}_t\cup\{(\mathcal{T}_k,a_k)\}$

\ENDFOR

\STATE $a_t \gets \textsc{Aggregate}(q_t,\{q_k^{sub}\}_t,\{a_k\}_t)$
\STATE \RETURN $a_t,h_t,\mathcal{G}_t$
\end{algorithmic}
\end{algorithm}
\section{Experiments}
\label{sec:exp}

\begin{table*}[t]
\centering
\scriptsize
\renewcommand{\arraystretch}{0.8}
\begin{tabular}{llccccc}
\toprule
\textbf{Reader} & \textbf{Method} & \textbf{Top-$k^\star$} & \textbf{Avg. Paras.} & \textbf{EM $\uparrow$ (\%)} & \textbf{F1 $\uparrow$ (\%)} & \textbf{GoldCtx $\uparrow$ (\%)} \\
\midrule
\multirow{21}{*}{Qwen3-32B}
& Direct C-RAG & 20 & 20 & 35.66 & 48.81 & 88.42 \\
& Direct C-RAG (With Thinking Mode) & 20 & 20 & 38.23 & 50.20 & 88.42 \\
\cmidrule(l){2-7}

& ReAct~\cite{yao2023react} & 20 & 20 & 23.70 & 37.80 & 90.70 \\     
& Self-Ask~\cite{press2023selfask} & 20 & 20 & 27.90 & 38.50 & 84.00 \\     
& HippoRAG~\cite{gutierrez2024hipporag} & 20 & 20 & 28.17 & 40.49 & 72.15 \\  
& SuRe~\cite{kim2024sure} & 20 & 20 & 30.30 & 43.00 & 89.70 \\  
& Adaptive-RAG~\cite{jeong2024adaptiverag} & 20 & 20 & 30.30 & 43.70 & 88.50 \\  
& IRCoT~\cite{trivedi2023ircot} & 20 & 20 & 33.10 & 46.20 & 88.80 \\     
\cmidrule(l){2-7}

& ChatQA~\cite{liu2024chatqa} & 20 & 20 & 34.93 & 47.19 & 86.14 \\     
& ConvSearch-R1~\cite{zhu2025convsearchr1} & 20 & 20 & 33.39 & 47.96 & \textbf{93.38} \\     
\cmidrule(l){2-7}

& RQ-RAG~\cite{chan2024rqrag} & 5 & 14.72 & 28.05 & 38.82 & 77.95 \\
& RQ-RAG$^\dagger$ & 5 & 14.72 & 31.26 & 42.34 & 77.95 \\
& ChainRAG$^\dagger$~\cite{zhu2025chainrag} & 20 & 21.76 & 35.54 & 48.72 & 83.88 \\
& LogicRAG$^\dagger$~\cite{chen2026logicrag} & 20 & 26.83 & 37.32 & 51.69 & 82.28 \\
\cmidrule(l){2-7}

& RLTST~\cite{fan2023rltst} & 20 & 20 & 32.20 & 44.70 & 89.60 \\     
& StructuredDDP~\cite{chi2022structureddp} & 20 & 20 & 36.30 & 50.80 & 85.60 \\     
\cmidrule(l){2-7}

\rowcolor{cyan!6!blue!4}
\cellcolor{white}  
& CMT-RAG (ours) & 5 & \textbf{13.99} & \textbf{41.73} & \textbf{55.63} & 86.25 \\
\rowcolor{cyan!6!blue!4}
\cellcolor{white} 
& Oracle traces & 5 & 14.90 & 42.18 & 56.23 & 88.08 \\
\midrule

\multirow{7}{*}{Llama-3.3-70B-Instruct}
& IRCoT~\cite{trivedi2023ircot} & 20 & 20 & 37.27 & 48.72 & 88.77 \\
& StructuredDDP~\cite{chi2022structureddp} & 20 & 20 & 37.81 & 49.31 & 85.56 \\
& ConvSearch-R1~\cite{zhu2025convsearchr1} & 20 & 20 & 40.57 & 52.57 & \textbf{93.88} \\
& LogicRAG$^\dagger$~\cite{chen2026logicrag} & 20 & 32.13 & 39.29 & 53.15 & 80.39 \\
& Direct C-RAG & 20 & 20 & 40.62 & 53.70 & 88.42 \\
\cmidrule(l){2-7}

\rowcolor{cyan!6!blue!4}
\cellcolor{white} 
& CMT-RAG (ours) & 5 & \textbf{14.05} & \textbf{44.70} & \textbf{57.55} & 85.10 \\
\rowcolor{cyan!6!blue!4}
\cellcolor{white} 
& Oracle traces & 5 & 15.30 & 47.22 & 60.82 & 89.83 \\
\bottomrule
\end{tabular}
\caption{Main results on the MuMu-QA long-dialogue split. All non-oracle systems use DRAGON retrieval. Top-$k^\star$ is selected from $k\!\in\!\{5,10,20\}$ by F1 for each baseline. Avg. Paras. denotes the mean number of unique paragraphs in the reader context per turn after deduplication, and GoldCtx the mean recall of gold supporting paragraphs. $\dagger$ denotes replaying the accumulated question--answer history before answer generation. Oracle traces replace only the generated trace drafts with gold traces, leaving retrieval and the reader unchanged.}
\label{tab:main_results}
\end{table*}

We evaluate whether complementary sub-question-level \emph{memory traces} improve multi-turn multi-hop conversational RAG. The experiments address five questions. \textbf{RQ1} Does CMT-RAG improve answer accuracy across different stateless readers and five categories of baseline methods? \textbf{RQ2} What gains come from the trace-management framework and the trained trace generator? \textbf{RQ3} How do recurrent SSM state and persistent DAG memory contribute across dialogue lengths? \textbf{RQ4} How does the trace-generator backbone affect answer quality and latency? \textbf{RQ5} Does CMT-RAG transfer to additional shared-corpus RAG benchmarks?

\subsection{Experimental Setup}

\paragraph{Dataset.}
As detailed in Appendix~\ref{app:splits-details}, MuMu-QA comprises three dialogue-length regimes: short (3--7 turns), long (6--32 turns), and ultra-long (33--67 turns). The short and long splits are used for training, while the ultra-long split is reserved for the evaluation of length-extrapolation.

\paragraph{Baselines.}
We compare CMT-RAG with five baseline families. Direct C-RAG retrieves with the unresolved current-turn query. Iterative and agentic retrieval methods include ReAct, Self-Ask, HippoRAG, SuRe, Adaptive-RAG, and IRCoT. Conversational context methods include ChatQA and ConvSearch-R1. Query-decomposition methods include RQ-RAG, ChainRAG, and LogicRAG. Dialogue-structure methods include RLTST and StructuredDDP.
Within each reader setting, all non-oracle systems use the same DRAGON corpus index, while each baseline retains its native reasoning or decomposition procedure. The $\dagger$ variants additionally replay the accumulated question--answer history at each turn.

\paragraph{Implementation details.}
We initialize the trace generator from Mamba-2-2.7B and train LoRA adapters with the three-stage SFT curriculum in \Cref{sec:joint-training}, followed by reader-specific DPO with reward weights $\gamma=0.2$. Unless otherwise specified, CMT-RAG retrieves five paragraphs per resolved sub-question using DRAGON and additional traces via keyword-overlap DAG lookup. Full training and hyperparameter details are provided in Appendix~\ref{app:trace-generator-details}.

\paragraph{Evaluation metrics.}
For end-task QA, we report Exact Match (EM) and token-level F1 against turn-level gold answers. Avg. Paras. is the mean number of unique paragraphs in the final reader context per turn after merging and deduplication. GoldCtx is the mean per-turn recall of gold supporting paragraphs in that final context. For efficiency analyses, end-to-end latency includes trace or plan generation, retrieval, intermediate reader calls, and final-answer generation, while excluding one-time model and index initialization. Further details appear in Appendix~\ref{app:evaluation-details}.

\subsection{Main Results across Readers (RQ1)}
We evaluate on the MuMu-QA long-dialogue split with Qwen3-32B and Llama-3.3-70B-Instruct as stateless readers. The two CMT-RAG variants share the same Stage~3 SFT checkpoint, while each uses a DPO adapter trained from preference pairs generated with the corresponding reader. This protocol evaluates compatibility with two readers rather than zero-shot reader swapping. At inference, CMT-RAG carries the SSM state across turns, retrieves fresh paragraphs with DRAGON~\cite{lin2023dragon} from resolved sub-questions, and uses trace keywords for long-range DAG lookup.

\Cref{tab:main_results} shows that the iterative baselines do not surpass Direct C-RAG under the shared evaluation protocol. ConvSearch-R1 obtains the highest GoldCtx recall without attaining the highest answer accuracy. CMT-RAG achieves the best non-oracle EM/F1 with Qwen3-32B (41.73/55.63) and Llama-3.3-70B-Instruct (44.70/57.55). Relative to Direct C-RAG, the gains are 6.07 EM and 6.82 F1 with Qwen and 4.08 EM and 3.85 F1 with Llama. These gains are obtained with approximately 14 unique paragraphs per turn instead of 20. CMT-RAG does not attain the highest GoldCtx recall, so the evidence supports more effective use of a smaller retrieved context rather than uniformly better supporting-paragraph retrieval. Oracle traces add 0.60 F1 with Qwen and 3.27 F1 with Llama, quantifying the remaining headroom in trace generation.

\subsection{Component Contributions (RQ2)}

To isolate the contributions of the CMT framework and the trained trace generator, we compare three settings. (1) A control setting that performs reader-based query decomposition without constructing memory traces or modeling explicit cross-turn dependencies. (2) A CMT-only setting that introduces the complete trace-management framework but uses the reader, instead of a trained SSM, to generate trace drafts. (3) The full CMT-RAG model, in which the trace drafts are generated by the SFT+DPO-trained trace generator.

\begin{table}[H]
\centering
\scriptsize
\renewcommand{\arraystretch}{0.8}
\setlength{\tabcolsep}{3.8pt}
\begin{tabular}{cccccccc}
\toprule
\multicolumn{2}{c}{\textbf{Components}}
&
\multicolumn{3}{c}{\textbf{Qwen3-32B}}
&
\multicolumn{3}{c}{\textbf{Llama-3.3-70B-It}}\\
\cmidrule(r){1-2}
\cmidrule(lr){3-5}
\cmidrule(l){6-8}
CMT
& L. Gen.
& MP
& EM $\uparrow$ (\%)
& F1 $\uparrow$ (\%)
& MP
& EM $\uparrow$ (\%)
& F1 $\uparrow$ (\%) \\
\midrule
\ding{53} & \ding{53}
& 9.79 & 35.67 & 47.11
& 9.89 & 39.38 & 51.72 \\

\ding{51} & \ding{53}
& 13.16 & 37.86 & 50.89
& 13.77 & 41.46 & 54.38 \\

\ding{51} & \ding{51}
& 13.99 & \textbf{41.73} & \textbf{55.63}
& 14.05 & \textbf{44.70} & \textbf{57.55} \\
\bottomrule
\end{tabular}
\caption{Ablation of the CMT framework and learned trace generator (L. Gen.). CMT performs trace construction, persistence, and DAG lookup. When L. Gen. is disabled, the reader generates trace drafts. MP denotes the mean number of unique paragraphs provided to the reader per turn.}
\label{tab:ablation_components}
\end{table}

\Cref{tab:ablation_components} separates the gain from complementary memory traces and the learned generator. With reader-generated traces, enabling CMT raises F1 by 3.78 points for Qwen and 2.66 points for Llama. Replacing reader-generated drafts with the trained SSM trace generator yields a further 4.74 and 3.17 F1-point gain, respectively. Although CMT increases the average number of retrieved paragraphs from about 10 to 14 per turn, the consistent improvements in EM and F1 indicate that the additional evidence is effectively utilized rather than introducing distracting context.

\subsection{Memory across Dialogue Lengths (RQ3)}
We conduct an ablation study to evaluate the two complementary memory components of CMT-RAG. All experiments follow the same settings as \Cref{tab:main_results} with Qwen3-32B. We further conduct evaluation on the ultra-long dialogue split with 33--67 turns of MuMu-QA to assess the robustness of the trained trace generator under dialogue lengths beyond those seen during training.

\begin{figure}[t]
  \centering
  \includegraphics[width=\columnwidth]{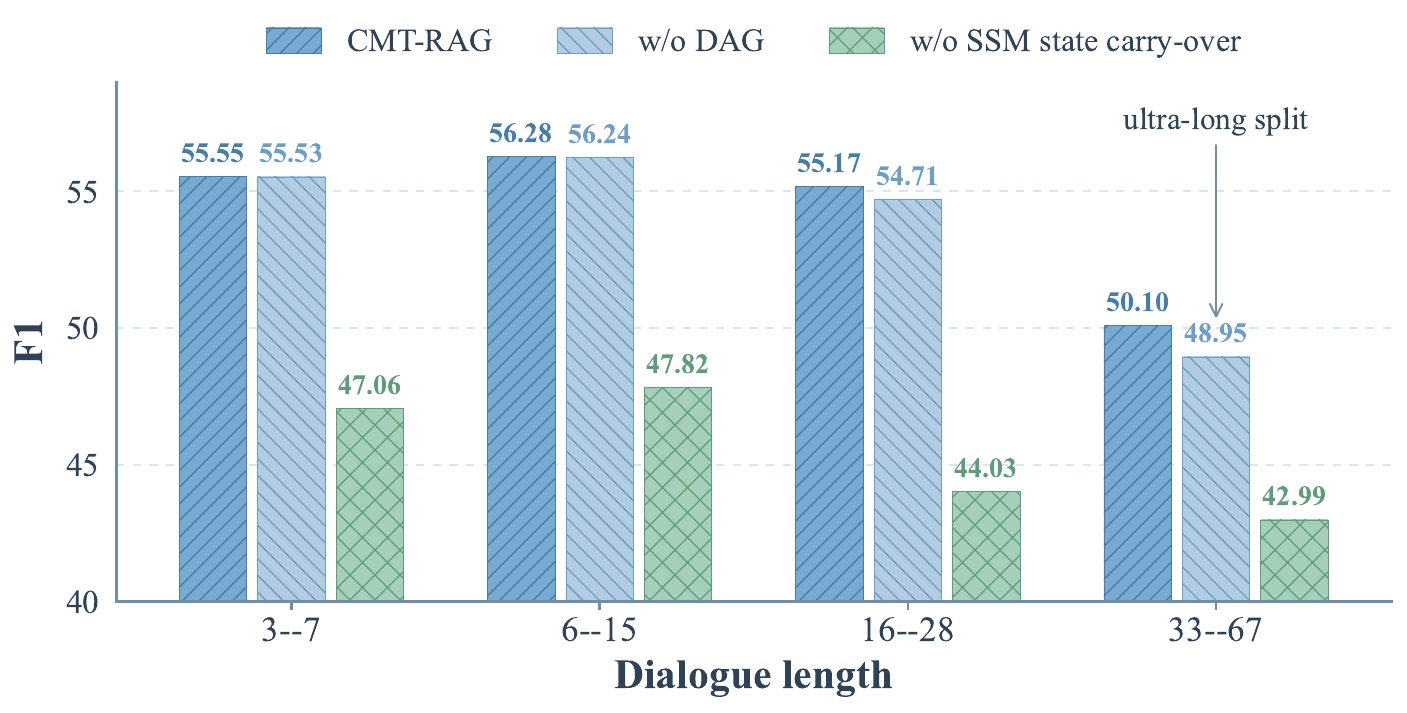}
  \caption{Ablation of the complementary memory framework across dialogue lengths. \emph{w/o DAG} removes persistent trace-DAG lookup while retaining SSM runtime memory. \emph{w/o SSM state} disables cross-turn SSM state carry-over while retaining persistent trace-DAG memory.}
  \label{fig-ablation}
\end{figure}

\Cref{fig-ablation} shows distinct length profiles for the two memory channels. Removing the DAG changes F1 by 0.02, 0.04, 0.46, and 1.15 points across the 3--7, 6--15, 16--28, and 33--67 turn bins, respectively, concentrating the DAG benefit in longer dialogues. Removing cross-turn SSM state reduces F1 by 8.49, 8.46, 11.14, and 7.11 points, making recurrent state the larger contributor in every length regime. Overall, the results confirm the complementary roles of the two memory channels across different dialogue lengths.

\subsection{Backbone Comparison (RQ4)}
We further evaluate a Transformer-based variant of CMT-RAG by replacing the original SSM backbone with Pythia-2.8B~\cite{biderman2023pythia}, whose parameter scale is comparable to that of the original Mamba2-2.7B backbone. We train the model on MuMu-QA using the same curriculum SFT and DPO procedure as the SSM-based counterpart. The resulting model serves as the decomposer and produces the same structured trace format described in \Cref{sec:ssm-decomposer}.

\begin{table}[H]
\centering
\scriptsize
\renewcommand{\arraystretch}{0.8}
\setlength{\tabcolsep}{5pt}
\begin{tabular}{llcccc}
    \toprule
    \textbf{Model} & \textbf{Group} & \textbf{EM $\uparrow$ (\%)} & \textbf{F1 $\uparrow$ (\%)} &
    \textbf{Avg. Paras.} & \textbf{E2E Time $\downarrow$ (s)} \\
    \midrule
    Trans. & SFT     & 39.96 & 53.83 & 11.93 & 1.59 \\
    \rowcolor{cyan!6!blue!4}
    SSM          & SFT     & \textbf{40.14} & \textbf{53.90} & 13.29 & \textbf{0.75} \\
    \midrule
    Trans. & +DPO & 40.09 & 54.03 & 11.96 & 1.61 \\
    \rowcolor{cyan!6!blue!4}
    SSM          & +DPO & \textbf{41.73} & \textbf{55.63} & 13.99 & \textbf{0.83} \\
    \bottomrule
\end{tabular}
\caption{Comparison of SSM and Transformer trace generators on the MuMu-QA long-dialogue split using Qwen3-32B as the reader. The SSM and Transformer are instantiated with Mamba-2-2.7B and Pythia-2.8B, respectively. E2E time (s/turn) includes decomposition, retrieval, and reader inference.}
\label{tab:ssm_transformer_comparison}
\end{table}

As shown in \Cref{tab:ssm_transformer_comparison}, the SSM-based decomposer consistently outperforms the Transformer baseline under both training settings. Under SFT, it achieves modest gains of 0.18 EM and 0.07 F1. The advantage becomes larger after DPO, reaching 41.73 EM and 55.63 F1, surpassing the Transformer by 1.64 EM and 1.60 F1, respectively. Beyond answer quality, the SSM is also substantially more efficient. In our implementation, it maintains a recurrent state across turns, whereas Pythia re-encodes the accumulated dialogue history at every turn, reducing end-to-end latency from 1.61 to 0.83 seconds per turn after DPO (a 48.4\% reduction). 

\begin{table*}[t]
\centering
\scriptsize
\renewcommand{\arraystretch}{0.8}
\setlength{\tabcolsep}{4pt}
\begin{tabular}{lcccccccccc}
\toprule
& \multicolumn{4}{c}{\textbf{RECOR}}
& \multicolumn{3}{c}{\textbf{HotpotQA}}
& \multicolumn{3}{c}{\textbf{2WikiMultiHopQA}} \\
\cmidrule(lr){2-5}
\cmidrule(lr){6-8}
\cmidrule(lr){9-11}
\textbf{Method}
& F1 $\uparrow$ (\%)
& BLEU-1 $\uparrow$
& ROUGE-L $\uparrow$
& E2E Time $\downarrow$ (ms)
& EM $\uparrow$ (\%)
& F1 $\uparrow$ (\%)
& E2E Time $\downarrow$ (ms)
& EM $\uparrow$ (\%)
& F1 $\uparrow$ (\%)
& E2E Time $\downarrow$ (ms) \\
\midrule
IRCoT
& 31.85 & 26.04 & 22.05 & 3192
& 33.02 & 42.39 & 2903
& 23.24 & 27.11 & 2971 \\
ConvSearch-R1
& 31.35 & 24.66 & 21.71 & 5648
& 27.99 & 39.45 & 3357
& 14.07 & 20.72 & 3751 \\
LogicRAG
& 28.98 & 22.58 & 19.97 & 14786
& \underline{43.65} & \textbf{56.95} & 4474
& \textbf{38.11} & \textbf{44.68} & 4794 \\
StructuredDDP
& \underline{32.80} & 26.88 & 22.66 & 1987
& -- & -- & --
& -- & -- & -- \\
Direct C-RAG
& 32.57 & \underline{29.70} & \underline{23.19} & 343
& 38.31 & 49.16 & 182
& 27.25 & 32.64 & 147 \\
\midrule
\rowcolor{cyan!6!blue!4}
CMT-RAG (ours)
& \textbf{36.42} & \textbf{33.52} & \textbf{25.70} & 2455
& \textbf{44.92} & \underline{56.76} & 549
& \underline{36.42} & \underline{42.93} & 570 \\
\bottomrule
\end{tabular}
\caption{Results on RECOR, HotpotQA, and 2WikiMultiHopQA. All methods retrieve from the same open corpus. We report token-level F1, BLEU-1, and ROUGE-L on RECOR; EM and token-level F1 on HotpotQA and 2WikiMultiHopQA. E2E time includes decomposition (or reasoning), passage retrieval, and reader inference, averaged over dialogue turns. All methods rerank the merged retrieval candidates and retain at most 10 passages for the final reader.}
\label{tab:hotpot_2wiki_recor_main}
\end{table*}

\subsection{Transfer across Benchmarks (RQ5)}
\label{sec:open-set_results}

We further evaluate transfer on the conversational retrieval benchmark RECOR~\cite{ali2026recor} and the single-turn multi-hop QA benchmarks HotpotQA~\cite{yang2018hotpotqa} and 2WikiMultiHopQA~\cite{ho2020twowikimultihopqa}, using Qwen3-32B as the reader. All methods retrieve from a shared benchmark corpus with the same DRAGON dense retriever~\cite{lin2023dragon}; for HotpotQA and 2WikiMultiHopQA, the released per-example contexts are merged into benchmark-level corpora, rather than using the official FullWiki setting. We report EM and token-level F1 on the two development sets, and F1, BLEU-1, and ROUGE-L across all RECOR dialogue turns.

\Cref{tab:hotpot_2wiki_recor_main} compares CMT-RAG with the strongest implemented representative from each baseline family. On the multi-turn benchmark RECOR, CMT-RAG achieves the best results on all reported metrics, demonstrating that complementary memory traces effectively capture and reuse conversational context beyond the setting of MuMu-QA. On the single-turn benchmarks HotpotQA and 2WikiMultiHopQA, CMT-RAG also remains competitive, achieving the best EM on HotpotQA and the second-best EM and F1 on 2WikiMultiHopQA. Although LogicRAG attains slightly stronger results on 2WikiMultiHopQA, it relies on multiple iterative reader calls, incurring substantially higher inference latency and requiring 8.4$\times$ more end-to-end inference time than CMT-RAG. By contrast, CMT-RAG delegates conversational parsing and memory management to a lightweight SSM trace generator, achieving a substantially better accuracy–efficiency trade-off while remaining competitive on single-turn multi-hop QA.
\section{Related Work}

Conversational RAG and multi-hop retrieval address complementary requirements of context-dependent information seeking. CMT-RAG connects these directions by maintaining retrieval-oriented sub-question traces whose dependencies and evidence persist across dialogue turns.

Conversational RAG methods represent dialogue context through history encoding~\cite{yang2025contextualretrieval, qian2022globalhistory}, memory~\cite{ye2026hmem, zhu2025cidgraphrag, zhong2024memorybank}, or query reformulation~\cite{wu2022conqrr,anantha2021qrecc}. ChatQA~\cite{liu2024chatqa} encodes conversational context for retrieval, and ConvGQR~\cite{mo2023convgqr} and ConvSearch-R1~\cite{zhu2025convsearchr1} reformulate context-dependent turns into standalone queries. These approaches effectively resolve local ambiguity and coreference, but leave retrieval dependencies among intermediate sub-questions implicit. Dialogue discourse parsers and graph-based conversational models make cross-turn structure explicit~\cite{li2020molweni,shi2019dialogue,fan2023rltst,chi2022structureddp}. Their nodes typically represent utterances or discourse units, and their edges encode discourse relations rather than dependencies between the sub-questions that drive evidence retrieval. CMT-RAG instead represents dialogue context at retrieval-oriented  granularity, allowing each current sub-question to address the specific prior trace and evidence on which it depends.

Reasoning-based retrieval decomposes complex questions into simpler units~\cite{huang2023qdt,wolfson2020break,perez2020unsupervised} or alternates retrieval with reasoning~\cite{asai2024selfrag, verma2024planrag}. QDG~\cite{hasson2021qdg}, RQ-RAG~\cite{chan2024rqrag}, ChainRAG~\cite{zhu2025chainrag}, and LogicRAG~\cite{chen2026logicrag} expose sub-question or dependency structure, whereas IRCoT~\cite{trivedi2023ircot} and Self-Ask~\cite{press2023selfask} generate intermediate reasoning steps that guide successive retrieval. These methods generally operate on self-contained queries and do not persist the resulting sub-questions, dependencies, and evidence as dialogue-level memory. Corpus-level graph RAG methods organize relations among documents or entities~\cite{gutierrez2024hipporag, edge2024graphrag}, which complements rather than captures conversational dependencies. CMT-RAG uses a session-level trace DAG whose nodes bind resolved sub-questions, dependency links, lookup keywords, and supporting evidence, making the retrieval unit and the memory unit the same persistent object across turns. 

\section{Conclusion}
We present \textbf{CMT-RAG}, a complementary memory framework that combines recurrent SSM state for local conversational context with a persistent trace DAG for long-range dependency resolution and evidence reuse. Each trace binds a resolved sub-question, dependency links, lookup keywords, and trace-local supporting evidence while keeping the reader stateless. We also introduce \textbf{MuMu-QA}, which provides sub-question-level cross-turn supervision and long-dialogue evaluation. Experiments demonstrate that CMT-RAG improves answer accuracy across two reader backbones while maintaining compact retrieval contexts. As CMT-RAG relies on accurate sub-question decomposition and dependency prediction, future work will focus on more robust trace generation. In addition, MuMu-QA is synthetic and inherits the domain bias of MuSiQue, motivating further evaluation on human-authored conversations.

\bibliography{custom}

\clearpage
\appendix
\section{MuMu-QA Construction}
\label{app:mumu-qa}

MuMu-QA is designed to evaluate multi-turn multi-hop RAG under sub-question-level cross-turn dependencies. Starting from the sub-question decompositions and supporting evidence provided by MuSiQue~\cite{trivedi2022musique}, we reorganize independent reasoning chains into conversational sessions in which later turns may depend on intermediate results established earlier. The construction process preserves the original reasoning and evidence supervision while introducing dialogue-level dependency structure, enabling controlled evaluation of sub-question decomposition, cross-turn trace linking, and evidence reuse. This section details the source data, dialogue synthesis procedure, split construction, and annotation schema.

\subsection{Source Data and Filtering}

MuMu-QA is constructed from the answerable split of MuSiQue, which provides multi-hop questions together with supporting paragraphs, sub-question decompositions, and intermediate answers. We exclude the unanswerable portion of the full split because MuMu-QA targets cross-turn dependency tracking rather than answerability detection or refusal behavior. We further remove near-duplicate examples whose decompositions and answers are effectively identical, preventing synthesized dialogues from collapsing into paraphrased repetitions of the same reasoning chain.

\subsection{Dialogue Synthesis}

MuMu-QA is synthesized in two stages. First, the graph operations construct sub-question nodes, intermediate answers, dependency edges, and evidence annotations directly from the MuSiQue reasoning graphs. Second, an LLM realizes the resulting graph fragments as natural conversational questions while preserving the underlying reasoning structure. Long- and ultra-long dialogues are subsequently obtained by interleaving synthesized sessions and globally remapping trace identifiers, dependencies, and trace-level paragraph identifiers. The subsequent interleaving and identifier-remapping stages require no additional LLM calls.

\paragraph{Graph synthesis.}

MuMu-QA uses two complementary graph-level synthesis operations. \textbf{Sub-question Relocation} relocates an independently answerable sub-question between dialogue turns to transform an in-turn dependency into a cross-turn dependency. The affected questions are rewritten accordingly while preserving the remaining reasoning graph and paragraph supervision. \textbf{Graph Splicing} chains multiple MuSiQue reasoning graphs by making a follow-up graph depend on an intermediate answer established in an earlier turn. For each follow-up, we retain the minimal subgraph required to derive its final answer and reconnect the selected entry node to the preceding trace. A teacher LLM then rewrites the selected reasoning graph into a natural conversational question, while all intermediate answers, dependency relations, and supporting-evidence annotations are inherited directly from the underlying MuSiQue graphs.

\paragraph{LLM-based question realization.}

Only the reader-facing turn questions are generated by an LLM. All sub-question nodes, intermediate answers, dependency edges, and evidence annotations are deterministically inherited from the original MuSiQue graphs. Depending on the synthesis operator and graph structure, different question-realization prompts are applied, as summarized in Table~\ref{tab:mumu_prompt_paths}. Long- and ultra-long dialogue synthesis does not invoke the LLM again; these stages only interleave previously synthesized dialogues and globally remap sub-question identifiers, answer references, dependency edges, and paragraph indices.

\begin{table}[h]
\centering
\small
\renewcommand{\arraystretch}{0.8}
\begin{tabular}{llc}
\toprule
\textbf{Operator} & \textbf{Case} & \textbf{Realization} \\
\midrule
GS  & Original graph        & Original \\
    & Partial graph         & Prompt A \\
    & Dependency follow-up  & Prompt C \\
\midrule
SQR & Carrier question      & Prompt B \\
    & Relocated follow-up   & Prompt C \\
    & Unchanged question    & Original \\
\bottomrule
\end{tabular}
\caption{Question realization under the two Stage~2 synthesis operators. GS and SQR denote Graph Splicing and Sub-question Relocation, respectively. ``Original'' indicates direct reuse of the original (or conversationalized) MuSiQue question without LLM generation.}
\label{tab:mumu_prompt_paths}
\end{table}

The default realization backend uses the open-source LLM (currently \texttt{gpt-oss-120b}~\cite{openai2025gptoss}) with temperature 0.2, a maximum of 220 generated tokens, and at most two generation attempts. Any comparable instruction-tuned LLM can be used. We choose GPT-OSS solely because it is open-source and reproducible.

\begin{table*}[t]
\centering
\small
\renewcommand{\arraystretch}{0.8}
\begin{tabular}{llrrrrrrr}
\toprule
Split & Part. & Dial. & Turns & Avg. T & SubQ & Avg. SQ & X-Edge & Avg. X-E \\
\midrule
\multirow{2}{*}{Short-dialogue}
  & Train & 3,104 & 3--7   & 3.96  & 3--25   & 8.80   & 1--6   & 1.84 \\
  & Dev   & 362   & 3--7   & 3.56  & 3--25   & 8.25   & 1--6   & 1.95 \\
\midrule
\multirow{2}{*}{Long-dialogue}
  & Train & 5,045 & 6--32 & 12.17 & 6--112 & 27.37 & 2--26 & 5.71 \\
  & Dev   & 548   & 6--28  & 9.85  & 8--101  & 22.90  & 2--23  & 5.88 \\
\midrule
Ultra-long stress
  & Test   & 200   & 33--67 & 52.93 & 50--235 & 122.05 & 8--49 & 32.23 \\
\bottomrule
\end{tabular}
\caption{Statistics of the MuMu-QA splits. Each row reports the number of dialogues together with the range and average of dialogue turns, global sub-questions, and cross-turn dependency edges. Cross-turn edges refer only to dependencies pointing to sub-questions in earlier turns.}
\label{tab:mumqa_split_stats}
\end{table*}

\paragraph{Prompt templates.}

Prompt~A is used only when the first graph-splice turn corresponds to a dependency closure ending at an intermediate MuSiQue node rather than a complete source question. Prompt~B realizes the carrier turn after relocating an independently answerable sub-question from a later reasoning graph while preserving the carrier's original final answer. Prompt~C is shared by graph-splice follow-up turns and relocated source turns. It requires the generated question to refer to the previous intermediate answer through an entity-type-compatible expression (e.g., ``that person'' or ``that city'') rather than explicitly mentioning the answer itself. The complete prompt templates are listed below.

\paragraph{Prompt A (Partial graph realization).} Fuse a partial reasoning graph into a natural parent question.

\begin{lstlisting}[basicstyle=\ttfamily\footnotesize,breaklines=true]
You are given a partial reasoning graph from MuSiQue.

Generate one natural parent question whose answer is the target answer.

Input:
- Original MuSiQue question
- Selected reasoning steps
- Target answer

Requirements:
- The question must be answerable using only the selected reasoning steps.
- Do not reveal the target answer.
- Return JSON:
{"question": "..."}
\end{lstlisting}

\paragraph{Prompt B (Carrier question realization).} Generate a carrier question that naturally preserves a relocated sub-question.

\begin{lstlisting}[basicstyle=\ttfamily\footnotesize,breaklines=true]
You are given a reasoning graph containing its original reasoning chain and one relocated auxiliary sub-question.

Generate one natural parent question whose final answer remains unchanged while naturally incorporating the auxiliary reasoning step.

Input:
- Original MuSiQue question
- Carrier reasoning graph
- Relocated sub-question
- Target answer

Return JSON:
{"question": "..."}
\end{lstlisting}

\paragraph{Prompt C (Dependency follow-up realization).} Generate a context-dependent follow-up question using implicit references.

\begin{lstlisting}[basicstyle=\ttfamily\footnotesize,breaklines=true]
You are given a reasoning graph whose first step depends on a previous conversational answer.

Generate one natural follow-up question using the specified reference phrase (e.g., "that city") instead of explicitly mentioning the previous answer.

Input:
- Previous answer
- Reference phrase
- Selected reasoning graph
- Original MuSiQue question
- Final answer

Requirements:
- Use the reference phrase.
- Do not reveal either the previous answer or the final answer.
- Return JSON:
{"question": "..."}
\end{lstlisting}

\paragraph{Generation validation.}

Generated questions are automatically validated before being included in MuMu-QA. We reject generations that omit required reference phrases, reveal bridge entities or final answers, violate entity-type constraints, contain malformed or repetitive wording, or exceed the prescribed length limit. As a result, LLM generation is restricted to the natural-language realization of reader-facing questions, while the trace graph, dependency structure, intermediate answers, and supporting-evidence annotations remain identical to those inherited from the underlying MuSiQue graphs.

\subsection{Splits and Leakage Control}
\label{app:splits-details}

We partition dialogues by grouped supporting-document titles and bridge entities rather than synthesized dialogue identifiers, preventing train and development splits from sharing nearly identical evidence configurations or intermediate answers under different surface forms. Supporting paragraphs inherited from MuSiQue serve as the gold evidence annotations. Table~\ref{tab:mumqa_split_stats} summarizes the resulting dataset statistics. For each dialogue, we report the number of turns ($T_d$), global sub-questions ($S_d$), and cross-turn dependency edges ($E_d$). Turns, SubQ, and X-Edge denote the minimum--maximum ranges within each split, whereas Avg.\ T, Avg.\ SQ, and Avg.\ X-E report the corresponding unweighted per-dialogue averages. Here, X-Edge counts individual dependency links from a current-turn sub-question to prerequisite sub-questions introduced in earlier turns.

An additional integrity audit found no overlap between the final short-dialogue train and development partitions in dialogue identifiers, source MuSiQue identifiers, exact normalized tuples of source question, answer, ordered decomposition, and dependency annotations, normalized final-step signatures, or exact question--answer pairs. Only three cross-split turn pairs exceeded a lexical Jaccard threshold of $0.82$; these pairs used similar surface templates but differed in resolved antecedents and supporting evidence, and none matched under the exact decomposition--dependency signature.

\subsection{Annotation Schema}

Each dialogue is annotated with a dialogue-wide \texttt{global\_subquestions} namespace, cross-turn dependency links, and one trace record for every node in the session DAG. Each trace record contains a trace identifier, a sub-question, DAG lookup keywords, predecessor trace identifiers, and supporting paragraph identifiers.

Each trace is further annotated with a keyword string for DAG lookup. Keywords are typically obtained by rule-based extraction from the sub-question, removing stop words while preserving entities and \texttt{[A$i$]} references. For reference-dependent turns lacking an explicit subject, the hidden referent is first resolved from the dialogue history (e.g., replacing ``that college'' with ``Aims Community College'') before keyword extraction and automatic validation.

This annotation schema enables evaluation of whether a system can decompose each user turn into sub-question-level retrieval units, connect them to prerequisite traces across turns, and ground each trace in reusable paragraph-level evidence.

\subsection{Breakdown by Synthesis Mode}

MuMu-QA is synthesized using two complementary operations with different structural characteristics. \emph{Sub-question Relocation} introduces cross-turn dependencies by relocating intermediate reasoning steps across conversational turns, whereas \emph{Graph Splicing} constructs longer reasoning chains by connecting multiple source reasoning graphs. In the long-dialogue split, Graph Splicing accounts for 75.39\% of evaluation turns and Sub-question Relocation for the remaining 24.61\%. The mixed Stage~3 setting therefore reflects the natural distribution of both synthesis modes in the final benchmark.

\begin{table}[H]
\centering
\small
\renewcommand{\arraystretch}{0.8}
\setlength{\tabcolsep}{4pt}
\begin{tabular}{lccc}
\toprule
\textbf{Mode} & \textbf{EM $\uparrow$ (\%)} & \textbf{F1 $\uparrow$ (\%)} & \textbf{Avg. Paras.} \\
\midrule
Sub-question Relocation & 39.98 & 55.55 & 15.76 \\
Graph Splicing            & 42.31 & 55.66 & 13.41 \\
Mixed (long-dialogue)         & 41.73 & 55.63 & 13.99 \\
\bottomrule
\end{tabular}
\caption{Performance of CMT-RAG across different MuMu-QA synthesis modes. ``Avg.\ Paras.'' denotes the average number of unique retrieved paragraphs per turn after deduplication.}
\label{tab:data_split_breakdown}
\end{table}

As shown in Table~\ref{tab:data_split_breakdown}, CMT-RAG performs consistently across the two synthesis modes. Graph Splicing achieves higher EM (42.31 vs.\ 39.98) while using fewer retrieved paragraphs (13.41 vs.\ 15.76), whereas both modes obtain nearly identical F1 scores. The mixed long-dialogue split closely matches the overall benchmark performance, indicating that CMT-RAG generalizes well across dialogue synthesis strategies with different cross-turn dependency structures.

\section{Details of Trace Generator}
\label{app:trace-generator-details}

\subsection{Architecture}
\label{app:network-details}

\begin{figure}[t]
  \centering
  \includegraphics[width=\columnwidth]{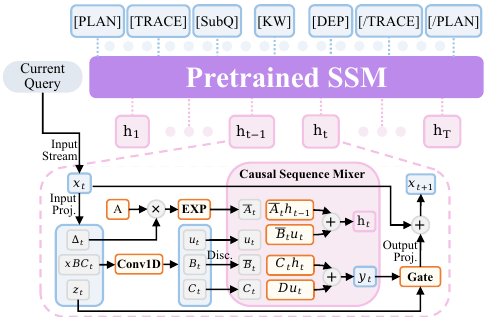}
  \caption{Architecture of the SSM-based trace generator. The recurrent SSM state carries dialogue context across turns, while the decoder generates structured trace drafts containing sub-question, keyword, and dependency fields.}
  \label{fig:module-details}
\end{figure}

Figure~\ref{fig:module-details} illustrates the architecture of the SSM-based trace generator. We instantiate the generator with a pretrained Mamba-2 backbone, whose selective state-space mixer maintains a recurrent hidden state throughout the dialogue. Unlike Transformer-based generators that must replay the entire dialogue history at every turn, the SSM processes each new query incrementally while propagating its hidden state between consecutive turns. This recurrent state serves as a compact short-term memory that summarizes recent conversational context and enables efficient long-dialogue generation.

At dialogue turn $t$, the current user query is appended to the input stream and processed together with the recurrent state $h_{t-1}$ inherited from the previous turn. After passing through the stacked selective SSM blocks, the model updates its hidden state to $h_t$, which is preserved and reused when processing the next user turn. During autoregressive decoding, the generator emits a structured trace set consisting of dedicated control tokens together with multiple trace fields,
\[
\begin{aligned}
&[\texttt{PLAN}]\\
&[\texttt{TRACE}],\;
[\texttt{SubQ}],\;
[\texttt{KW}],\;
[\texttt{DEP}],\;
[\texttt{/TRACE}]\\
&\cdots\\
&[\texttt{/PLAN}]
\end{aligned}
\]
where each \texttt{TRACE} record stores a decomposed sub-question together with its retrieval keywords and dependencies on prerequisite traces. These control and field tokens are added to the tokenizer vocabulary before fine-tuning so that the model can generate structured trace plans through standard causal language modeling.

Following common parameter-efficient fine-tuning practice, LoRA adapters are inserted only into the input and output projection layers of each Mamba-2 block, while all pretrained backbone parameters remain frozen. Consequently, the model learns to generate structured traces while preserving the long-context modeling capability inherited from the pretrained SSM.

\subsection{Computing Infrastructure}

All training and evaluation experiments were conducted on a Linux server equipped with four NVIDIA H100 GPUs (80\,GB HBM3 memory each; 320\,GB total), two Intel Xeon Platinum 8468V CPUs (96 physical cores and 192 hardware threads in total), and 2.0\,TiB system memory. The server ran Ubuntu 22.04.4 LTS with Linux kernel 5.14.0.

\paragraph{Training environment:}
Python 3.10.19;
PyTorch 2.3.1 (CUDA 12.1);
Transformers 4.43.0;
PEFT 0.11.1;
Mamba-SSM 2.2.2;
causal-conv1d 1.4.0;
Triton 2.3.1;
FAISS 1.8.0;
Sentence-Transformers 5.3.0.

\paragraph{Inference environment:}
We used Qwen3-32B as the primary reader and Llama-3.3-70B-Instruct for additional reader-backbone experiments. Both readers were served in BF16 using vLLM~0.11.0, PyTorch~2.8.0 with CUDA~12.8, and Transformers~4.57.1.

\subsection{Trace Generator Training}
\label{app:training-details}

Following the architecture described above, we optimize the trace generator using a three-stage supervised curriculum followed by Direct Preference Optimization (DPO). We use the GPT-NeoX tokenizer associated with Mamba-2-2.7B and extend its vocabulary with seven structural tokens:
\texttt{[PLAN]}, \texttt{[/PLAN]}, \texttt{[TRACE]}, \texttt{[/TRACE]}, \texttt{[SubQ]}, \texttt{[KW]}, and \texttt{[DEP]}.
Each generated trace consists of a decomposed sub-question, a retrieval keyword for DAG lookup, and a list of predecessor trace identifiers. During fine-tuning, we insert Low-Rank Adaptation (LoRA) modules into the Mamba \texttt{in\_proj} and \texttt{out\_proj} projections. All remaining backbone parameters remain frozen.

\paragraph{Curriculum Supervised Fine-Tuning.}

We optimize the trace generator using a three-stage supervised curriculum. Stage~1 trains a rank-16 LoRA adapter on 9,653 single-turn examples with 1,220 validation examples to learn the trace syntax and basic decomposition structure. The resulting adapter is merged into the backbone before Stage~2, which trains a rank-8 LoRA adapter on 3,104 multi-turn dialogues from the \texttt{short-dialogue} split. Stage3 continues training the same adapter on 5,045 dialogues from the \texttt{long-dialogue} training split, supplemented with 1,009 randomly sampled dialogues from the Stage2 training set for replay. Stage~2 and Stage~3 checkpoints are selected solely by the autoregressive validation loss over the linearized gold trace sequence. This criterion does not involve downstream retrieval or question-answering execution. Table~\ref{tab:ssm-training-config} summarizes the training hyperparameters.

\begin{table}[h]
\centering
\scriptsize
\renewcommand{\arraystretch}{0.8}
\setlength{\tabcolsep}{3.2pt}
\begin{tabular}{lcccc}
\toprule
Setting & Stage~1 SFT & Stage~2 SFT & Stage~3 SFT & DPO \\
\midrule
Training items & 9,653 & 3,104 & 6,054 & 45,420 pairs \\
Max input Seq. Len. & 1,024 & 4,096 & 8,192 & 5,120 \\
LoRA learning rate
& $5\!\times\!10^{-4}$
& $5\!\times\!10^{-5}$
& $2\!\times\!10^{-5}$
& $5\!\times\!10^{-7}$ \\
Epochs & 3 & 3 & 3 & 1 \\
Batch / accumulation & 4 / 1 & 1 / 8 & 1 / 8 & 1 / 1 \\
LoRA rank / $\alpha$ & 16 / 32 & 8 / 16 & 8 / 16 & 8 / 16 \\
LoRA dropout & 0.05 & 0.10 & 0.10 & 0.00 \\
DPO $\beta$ & -- & -- & -- & 0.05 \\
AdamW weight decay & 0.05 & 0.05 & 0.05 & 0.01 \\
Training arithmetic & FP16 & FP16 & FP16 & FP32 \\
\bottomrule
\end{tabular}
\caption{Training configuration of the SSM trace generator used for the reported results. Training-item counts denote single-turn examples, dialogue sessions, or offline preference pairs. Stage~3 continues training from the Stage~2 LoRA adapter.}
\label{tab:ssm-training-config}
\end{table}

Unless otherwise specified, all SFT stages use AdamW with gradient clipping (maximum norm 1.0), a linear warmup over the first 5\% of optimization steps followed by linear learning-rate decay, and random seed 42.

\paragraph{Preference Optimization.}

Preference pairs are constructed from the selected Stage-3 SFT policy. All candidate traces and preference pairs are generated exclusively from the long-dialogue training split; no dialogue from the long-dialogue development split or the ultra-long test split is used for candidate sampling, reward computation, pair construction, or DPO training. For each training dialogue turn with context $c$, we sample four candidate traces using temperature $1.0$ and top-$k$ sampling ($k=40$). Each candidate trace $\tau$ is executed through the fixed retrieval–reader pipeline and assigned the reward

\begin{equation}
\label{eq:dpo-reward}
R(\tau;c)
=
F_{\mathrm{final}}(\tau)
+
\gamma F_{\mathrm{sub}}(\tau).
\end{equation}

\noindent
where $F_{\mathrm{final}}$ is the final-answer F1 and $F_{\mathrm{sub}}$ is the average F1 over lexical matched intermediate sub-questions. Intermediate sub-questions are matched to the reference decomposition using a minimum semantic-similarity threshold of 0.55. Preference pairs are formed from the highest- and lowest-reward candidates whenever their reward difference is at least 0.05. Candidate traces with invalid formats are filtered before finalizing the preference pairs, resulting in 45,420 offline preference pairs.

We merge the Stage~3 LoRA adapter into the backbone and attach a new rank-8 LoRA adapter containing 10.7M trainable parameters for DPO, while the merged Stage~3 policy serves as the frozen reference model. The policy is optimized for one epoch using standard DPO with $\beta=0.05$, batch size 1, and no auxiliary SFT loss. Notably, DPO is run for a pre-specified single epoch, without downstream-QA-based early stopping or checkpoint search.

\subsection{Ablation and Sensitivity Analysis of the DPO Reward}
\paragraph{Protocol.} The DPO reward in Equation~\ref{eq:dpo-reward} combines final-answer F1, $F_{\mathrm{final}}$, with an auxiliary matched sub-question score, $F_{\mathrm{sub}}$, weighted by $\gamma$. We use $\gamma=0.2$ as a single canonical coefficient in all main experiments. This value was fixed before conducting the sensitivity analysis below and is kept unchanged across reader backbones and downstream benchmark evaluations. Accordingly, the purpose of this analysis is not to select an optimal value of $\gamma$ on the development split, but to isolate the contribution of the auxiliary reward term relative to $\gamma=0$ and assess whether the resulting performance is robust across a broad range of positive coefficients.

\paragraph{Controlled comparison.} Starting from the same Stage-3 SFT checkpoint, we reuse the same candidate-trace pool sampled from the \texttt{long-dialogue} training split and vary only $\gamma\in\{0,0.2,0.4,0.6,0.8,1.0\}$. For each value, we recompute the composite rewards and reconstruct preference pairs using the same candidate-ranking procedure, reward-margin threshold, and validity-filtering criteria. The reader, model initialization, optimizer configuration, DPO epoch count, and all remaining evaluation settings are held fixed. Because changing $\gamma$ can alter both the candidate ordering and the set of pairs satisfying the reward-margin criterion, the exact preference-pair composition may differ across settings. This experiment therefore measures the sensitivity of the complete preference-construction procedure to $\gamma$, rather than reweighting a fixed set of preference pairs. Because the resulting models are compared on the MuMu-QA \texttt{long-dialogue} development split, we report the experiment as a controlled reward-component ablation and sensitivity analysis rather than as a held-out estimate of hyperparameter optimality.

\begin{figure}[t]
\centering
\includegraphics[width=0.94\linewidth]{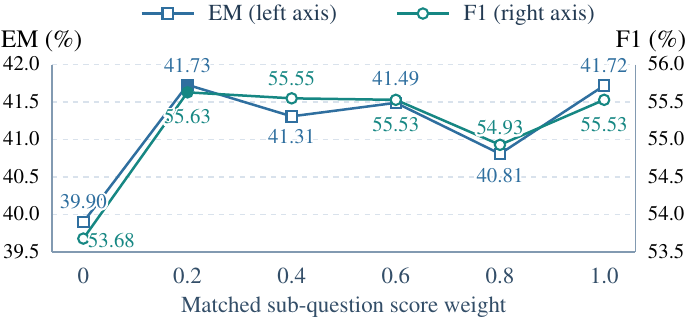}
\caption{Ablation and sensitivity analysis of the matched sub-question reward coefficient \(\gamma\) used for DPO preference construction on the MuMu-QA \texttt{long-dialogue} development split. EM is shown on the left axis and F1 on the right axis. The shaded band marks the fixed canonical setting \(\gamma=0.2\), which is used unchanged in all main DPO experiments; the remaining values are evaluated only in this controlled analysis.}
\label{fig:dpo-reward-weight-sensitivity}
\end{figure} 

\paragraph{Results.} As shown in Figure~\ref{fig:dpo-reward-weight-sensitivity}, removing the matched sub-question term by setting $\gamma=0$ yields EM/F1 scores of 39.90/53.68. The positive settings $\gamma=0.2$, $0.4$, $0.6$, $0.8$, and $1.0$ yield 41.73/55.63, 41.31/55.55, 41.49/55.53, 40.81/54.93, and 41.72/55.53, respectively. Every evaluated positive coefficient improves over $\gamma=0$ in the observed results, with gains of 0.91--1.83 EM points and 1.25--1.95 F1 points. Across all positive settings, EM spans 0.92 points and F1 spans 0.70 points. Moreover, four coefficients distributed across the evaluated range, $\gamma\in\{0.2,0.4,0.6,1.0\}$, are tightly clustered: their F1 scores span only 0.10 points, from 55.53 to 55.63, while their EM scores span 0.42 points, from 41.31 to 41.73. Although $\gamma=0.8$ produces a local decrease to 40.81/54.93, performance recovers to 41.72/55.53 at $\gamma=1.0$. The observed sensitivity is therefore non-monotonic and does not indicate systematic degradation as the auxiliary reward weight increases.

\paragraph{Interpretation.} The dominant empirical contrast is between removing and including the matched sub-question reward, rather than between individual positive coefficients. All evaluated positive weights outperform $\gamma=0$ in observed EM and F1, while several substantially different positive settings remain closely competitive. In particular, $\gamma=1.0$, which assigns equal nominal coefficients to the two normalized reward components, performs nearly identically to the canonical setting $\gamma=0.2$. The analysis therefore supports the matched sub-question score as a useful additional preference signal, while providing no evidence of a narrow or uniquely optimal coefficient. Although $\gamma=0.2$ attains the highest observed EM and F1, the small differences among the competitive positive-weight settings are not interpreted as evidence that $\gamma=0.2$ is statistically superior. We retain $\gamma=0.2$ because it was fixed before this analysis and provides a conservative canonical configuration: since both reward components are normalized to $[0,1]$, it limits the maximum nominal contribution of the auxiliary term to 0.2 relative to 1.0 for the primary final-answer term, thereby preserving final-answer quality as the dominant reward component. No reader-specific or benchmark-specific retuning is performed.

\section{Inference and Evaluation}
\label{app:evaluation-details}




\subsection{Retrieval Details}

Cross-turn dependencies are first resolved using the dependency identifiers stored in each generated trace. The referenced traces are retrieved from the session DAG through their global namespace identifiers, and their intermediate answers are substituted into the corresponding placeholders to obtain resolved sub-questions. For the MuMu-QA experiments, each resolved sub-question retrieves its top-$5$ supporting paragraphs using the DRAGON dense retriever~\cite{lin2023dragon}. Freshly retrieved and trace-reused paragraphs are merged and deduplicated by paragraph identifier in first-occurrence order to form the reader context.

In parallel with explicit dependency access, lexical lookup searches completed same-dialogue trace nodes older than the preceding ten-turn recent window. Each keyword string is lowercased and tokenized into a set using the Python Unicode regular expression \texttt{\textbackslash w+}. For a generated keyword token set $Q$ and a historical keyword token set $K$, similarity is computed using the binary cosine, equivalently the Ochiai coefficient,
\[
s(Q,K)=
\begin{cases}
\dfrac{|Q\cap K|}{\sqrt{|Q||K|}}, & |Q|>0 \text{ and } |K|>0,\\
0, & \text{otherwise}.
\end{cases}
\]
The runtime matcher applies no stemming, additional stop-word removal, or frequency-based weighting. Traces already accessed through recent explicit dependencies are excluded from the lookup candidates, which are ranked by decreasing similarity score, with exact ties broken by increasing integer trace identifier. Only the highest-ranked candidate satisfying the strict threshold $s>0.05$ is retained. Its stored paragraph identifiers are used to recover the associated supporting paragraphs, which are merged with the DRAGON results and deduplicated by paragraph identifier.

\begin{figure*}[t]
  \centering
  \includegraphics[width=0.94\textwidth]{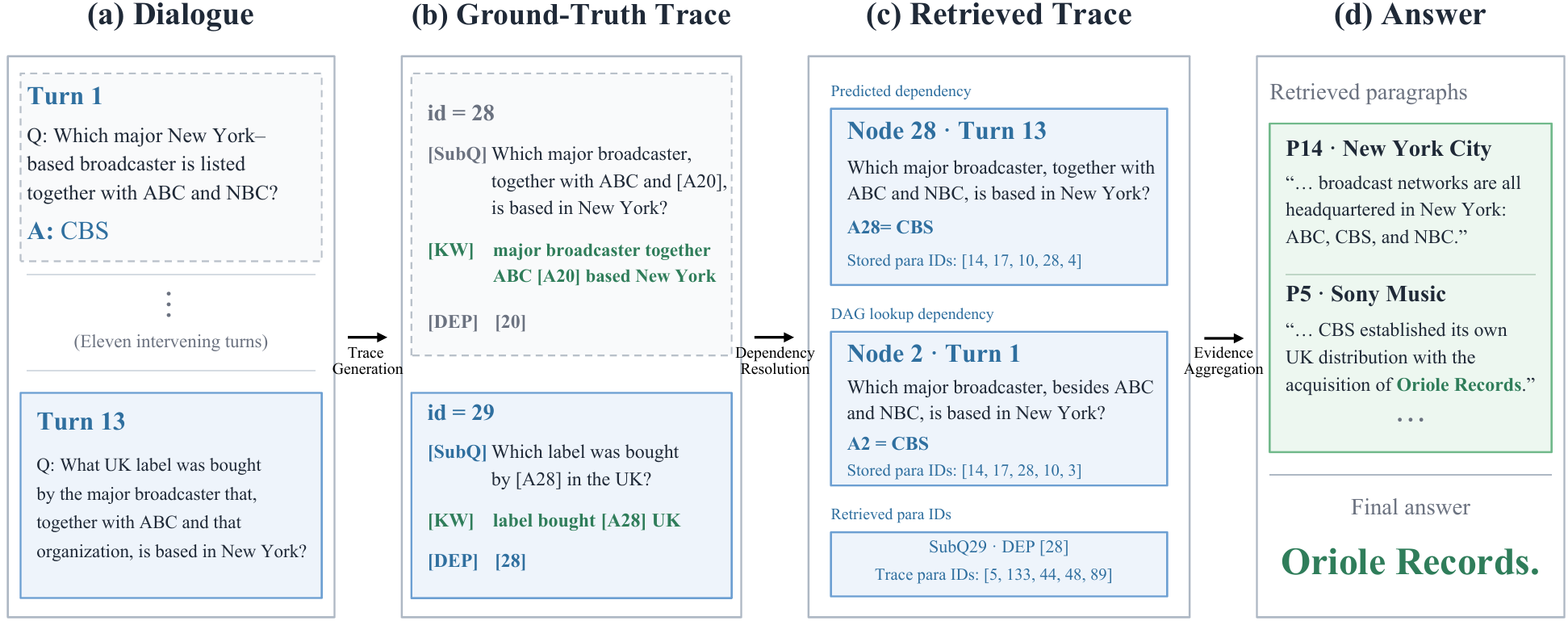}
  \caption{A real evaluation example from MuMu-QA. The current dialogue turn is first decomposed into structured traces with explicit dependencies and keywords. Dependency resolution combines the predicted dependency with a keyword-recovered historical trace node from the DAG, enabling evidence reuse across distant turns. The retrieved and reused evidence is then aggregated and passed to the stateless reader to produce the final answer.}
  \label{fig:trace-workflow-example}
\end{figure*}

\subsection{Reader Details}

The reader receives two types of inputs: (i) the resolved sub-question and (ii) the merged paragraphs retrieved for the current turn. It is instructed to answer strictly according to the retrieved evidence and to output only the shortest answer span (or \texttt{yes}/\texttt{no} when appropriate), without explanation.

\paragraph{System prompt.}
The system prompt defines the reader's role and output constraints.

\begin{lstlisting}[basicstyle=\ttfamily\footnotesize,breaklines=true]
Answer the question using only the provided paragraphs. Reply with only the answer span, yes, or no. Do not explain.
\end{lstlisting}

\paragraph{Sub-question prompt.}
Each resolved sub-question is answered using the following prompt template.

\begin{lstlisting}[basicstyle=\ttfamily\footnotesize,breaklines=true]
Provided paragraphs:
[Paragraph 1 | score={score_1}]
Title: {title_1}
{paragraph_text_1}

[Paragraph 2 | score={score_2}]
Title: {title_2}
{paragraph_text_2}

...

Question: {resolved sub-question}

Answer:
\end{lstlisting}

The final answer is aggregated using an analogous prompt but only includes the user query and reasoning steps from the current turn. Unless otherwise stated, inference uses a batch size of 8, at most 32 concurrent requests, and a maximum generation length of 512 tokens. For models supporting an explicit thinking mode, we disable it during both sub-question answering and final-answer generation.

\subsection{Stateful SSM Inference}

The trace generator performs stateful inference by preserving its recurrent state across dialogue turns, allowing each new turn to process only the current prompt instead of replaying the full dialogue history. Inference is performed in BF16 and greedy decoding with up to 512 generated tokens per turn. To support efficient multi-session serving, we batch up to 16 concurrent sessions and employ packed recurrent caches, GPU-resident decoding buffers, and CUDA Graph execution to reduce memory overhead and decoding latency.

\begin{table*}[t]
\centering
\scriptsize
\renewcommand{\arraystretch}{0.8}
\setlength{\tabcolsep}{6pt}
\begin{tabular}{lccccccc}
\toprule
\multicolumn{2}{c}{\textbf{Configuration}}
&
\multicolumn{3}{c}{\textbf{Qwen3-32B}}
&
\multicolumn{3}{c}{\textbf{Llama-3.3-70B-Instruct}}\\
\cmidrule(r){1-2}
\cmidrule(lr){3-5}
\cmidrule(l){6-8}
Method
& Top-$k$
& Avg. Paras.
& EM $\uparrow$ (\%)
& F1 $\uparrow$ (\%)
& Avg. Paras.
& EM $\uparrow$ (\%)
& F1 $\uparrow$ (\%) \\
\midrule
Direct C-RAG
& 20
& 20.00 & 32.15 & 44.98
& 20.00 & 36.60 & 48.74 \\

ConvSearch-R1~\cite{zhu2025convsearchr1}
& 20
& 20.00 & 30.80 & 44.30
& 20.00 & 38.70 & 49.70 \\

StructuredDDP~\cite{chi2022structureddp}
& 20
& 20.00 & 32.10 & 46.50
& 20.00 & 34.50 & 44.60 \\

IRCoT~\cite{trivedi2023ircot}
& 20
& 19.98 & 30.70 & 42.50
& 19.98 & 33.00 & 43.60 \\

LogicRAG~\cite{chen2026logicrag}
& 20
& 28.18 & 33.54 & 47.38
& 34.91 & 38.09 & 50.61 \\
\midrule
\textsc{CMT-RAG} (ours)
& 5
& 17.60 & \textbf{36.84} & \textbf{50.10}
& 15.73 & \textbf{40.11} & \textbf{51.78} \\
\bottomrule
\end{tabular}
\caption{Results on the ultra-long-dialogue split of MuMu-QA.}
\label{tab:ultra-long_qwen_llama_comparison}
\end{table*}

\subsection{Case Study}
\label{sec:trace-case-study}

\Cref{fig:trace-workflow-example} illustrates an evaluation example from MuMu-QA. At Turn~13, the user asks which UK label was acquired by the major broadcaster that, together with ABC and NBC, is based in New York. The trace generator decomposes the query into two traces. Trace~28 resolves dependency~20 to identify the broadcaster while generating retrieval keywords, and Trace~29 asks which UK label was acquired by the resolved entity through dependency~28.

Trace~28 resolves dependency~20 and identifies the broadcaster as \emph{CBS}. Using the generated keywords, the trace DAG retrieves an earlier node (Node~2) from Turn~1 and reuses its stored supporting paragraphs as complementary evidence. After Trace~28 is completed, the retriever obtains additional paragraphs for Trace~29, and the reader answers using the combined reused and newly retrieved evidence. The final answer is correctly identified as \emph{Oriole Records}. This example illustrates how CMT-RAG combines complementary memories: the recurrent trace resolves recent conversational dependencies, while the trace DAG supplies reusable long-range evidence beyond the local dialogue context.



\subsection{Evaluation and Configuration-Selection Protocol}

\paragraph{Evaluation split and default configuration.}
Tables~1--3 report results on the complete MuMu-QA \texttt{long-dialogue} development split, comprising 548 dialogues and 5,396 turns. Unless otherwise specified, CMT-RAG uses greedy trace generation, cross-turn SSM-state carry-over, DRAGON retrieval with \(k_{\mathrm{local}}=5\) per resolved sub-question, dependency-based DAG access, and at most one additional trace retrieved via keyword-overlap lookup. We report exact match (EM), token-level F1, GoldCtx---the mean per-turn recall of gold supporting paragraphs---and Avg. Paras., the average number of unique paragraphs provided to the reader after evidence merging and deduplication. End-to-end latency includes trace generation, retrieval, intermediate reader calls, and final-answer generation, but excludes one-time model and retriever initialization.

\paragraph{CMT-RAG configuration and checkpoint selection.}
For CMT-RAG and the Oracle-trace variant, the local DRAGON retrieval depth is fixed at \(k_{\mathrm{local}}=5\) per resolved sub-question in all MuMu-QA experiments; no development-set tuning is performed. Since each turn may generate multiple sub-questions and retrieval calls, the total retrieval budget depends on both \(k_{\mathrm{local}}\) and the number of generated traces. We therefore report Avg. Paras. as the realized reader-context size instead of comparing nominal per-call \(k\) across methods.

Stage-2 and Stage-3 SFT checkpoints are selected solely by the autoregressive validation loss on the linearized gold trace targets. End-task QA metrics (EM/F1), retrieval quality (GoldCtx), and efficiency are never used for checkpoint selection. DPO is trained for one pre-specified epoch, and the resulting final checkpoint is used without downstream-QA-based early stopping or checkpoint search.

\paragraph{Baseline retrieval-depth selection.}
For baseline methods only, we evaluate $k\in\{5,10,20\}$ and report the configuration with the highest final-answer F1 on the \texttt{long-dialogue} development split. This procedure reports the best observed development-set retrieval configuration for each baseline and is not applied to CMT-RAG or the Oracle-trace variant. For clarity, the superscript $\star$ in the Top-$k$ column of Table~1 denotes retrieval-depth selection for baseline rows only; it does not indicate a sweep for CMT-RAG or Oracle traces.

\paragraph{Interpretation of development-set results.}
Because the Stage-3 checkpoint is selected using trace-generation validation loss on the \texttt{long-dialogue} development split, and baseline retrieval depth is also selected on this split, the reported long-dialogue answer metrics should be interpreted as development-set comparisons rather than as estimates from an independently held-out IID test split. The \texttt{ultra-long} split is not used for training, checkpoint selection, or retrieval-depth selection and is used only for length-extrapolation evaluation.

\subsection{Corpus-level RAG Evaluation}

We evaluate on three corpus-level RAG benchmarks: HotpotQA~\cite{yang2018hotpotqa}, 2WikiMultiHopQA~\cite{ho2020twowikimultihopqa}, and RECOR~\cite{ali2026recor}. HotpotQA contains 7,405 development questions requiring multi-hop reasoning through bridge and comparison relations over two supporting Wikipedia documents. 2WikiMultiHopQA contains 12,576 development questions covering bridge, comparison, compositional, and inference reasoning over two to four Wikipedia documents. RECOR is a conversational retrieval benchmark comprising 707 multi-turn conversations and 2,971 evaluation turns across 11 domains, where each turn requires resolving dialogue context before retrieving supporting evidence.

For all three benchmarks, every method retrieves from a shared corpus rather than an example-specific context. For HotpotQA and 2WikiMultiHopQA, we construct the retrieval corpus by pooling all released contexts, yielding 507,494 and 398,354 title documents, respectively. Consequently, our HotpotQA protocol differs from the official FullWiki setting. For RECOR, we directly use the released corpus for each domain. Unless otherwise specified, Qwen3-32B serves as the reader. We report EM and token-level F1 on the HotpotQA and 2WikiMultiHopQA development sets, and token-level F1, BLEU-1, and ROUGE-L on all RECOR evaluation turns.

\subsection{Comparison on Ultra-long Stress Test Set}

To evaluate length extrapolation, we further compare CMT-RAG against representative conversational and multi-hop RAG baselines on the ultra-long MuMu-QA split (33–67 turns), which is excluded from training. Following the long-dialogue setting, all baselines retrieve up to 20 passages, while CMT-RAG retrieves five paragraphs per resolved sub-question with at most one additional trace from the DAG. 

As shown in \Cref{tab:ultra-long_qwen_llama_comparison}, CMT-RAG consistently achieves the best EM and F1 with both Qwen3 and Llama-3.3, while using substantially fewer retrieved paragraphs than the baselines. These results demonstrate that complementary memory traces effectively preserve reusable conversational context and generalize beyond the dialogue lengths seen during training.



\end{document}